\documentclass[twoside,11pt]{article}

\usepackage[preprint, nohyperref]{jmlr2e}
\usepackage{booktabs}

\usepackage{hyperref}
\usepackage[table,xcdraw]{xcolor}
\definecolor{bluecite}{HTML}{0875b7}
\hypersetup{
  colorlinks=true,
  citecolor=bluecite,
  linkcolor=magenta
}
\usepackage[export]{adjustbox}

\definecolor{mava}{HTML}{ec6e4d}
\definecolor{ogmarl}{HTML}{7751f3}
\definecolor{jumanji}{HTML}{3eb266}
\definecolor{flashbax}{HTML}{f0b13d}
\definecolor{marleval}{HTML}{6F3132}

\newcommand{\mava}{\textbf{\color{mava}{Mava}}}

\newcommand{\ogmarl}{\textbf{\color{ogmarl}{OG-MARL}}}
\newcommand{\jumanji}{\textbf{\color{jumanji}{Jumanji}}}
\newcommand{\flashbax}{\textbf{\color{flashbax}{Flashbax}}}
\newcommand{\marleval}{\textbf{\color{marleval}{MARL-eval}}}

\usepackage{xspace}
\usepackage{units}

\usepackage{microtype}
\usepackage{caption} 
\usepackage{subcaption}
\usepackage{comment}

\usepackage[parfill]{parskip}

\ShortHeadings{Mava: distributed  multi-agent reinforcement learning in JAX}{de Kock et al.}
\firstpageno{1}

\begin{document}

\title{Mava: a research library for distributed multi-agent reinforcement learning in JAX}

\author{\name Ruan de Kock\thanks{Equal contribution. Correspondence: r.dekock@instadeep.com} 
       \name \hspace{1em} Omayma Mahjoub$^{*}$ 
       \name \hspace{1em} Sasha Abramowitz$^{*}$ 
       \name \hspace{1em} Wiem Khlifi \email \vspace{0.5em}  \\ 
       \name Callum Rhys Tilbury
       \name \hspace{1em} Claude Formanek 
       \name \hspace{1em} Andries Smit 
       \name \hspace{1em} Arnu Pretorius \\
       \vspace{-1em}
       \AND
       \addr \hspace{18.5em} InstaDeep\\
       \addr  \quad
}

\editor{xxx}

\maketitle
\thispagestyle{empty}

\begin{abstract}
Multi-agent reinforcement learning (MARL) research is inherently computationally expensive and it is often difficult to obtain a sufficient number of experiment samples to test hypotheses and make robust statistical claims. Furthermore, MARL algorithms are typically complex in their design and can be tricky to implement correctly. These aspects of MARL present a difficult challenge when it comes to creating useful software for advanced research. Our criteria for such software is that it should be simple enough to use to implement new ideas quickly, while at the same time be scalable and fast enough to test those ideas in a reasonable amount of time. In this preliminary technical report, we introduce Mava, a research library for MARL written purely in JAX, that aims to fulfill these criteria. We discuss the design and core features of Mava, and demonstrate its use and performance across a variety of environments. In particular, we show Mava's substantial speed advantage, with improvements of 10-100x compared to other popular MARL frameworks, while maintaining strong performance. This allows for researchers to test ideas in a few minutes instead of several hours. Finally, Mava forms part of an ecosystem of libraries that seamlessly integrate with each other to help facilitate advanced research in MARL. We hope Mava will benefit the community and help drive scientifically sound and statistically robust research in the field. The open-source repository for Mava is available at \href{https://github.com/instadeepai/Mava}{https://github.com/instadeepai/Mava}.

\end{abstract}




\section{Introduction}
To facilitate advanced research in MARL from an engineering perspective presents a difficult challenge: to strike a balance between code that is highly performant, to run computationally expensive experiments efficiently, and code that is easy to understand, to enable conceptual and algorithmic development in a reasonable time frame. Software that is very simple to understand may likely be too slow to pursue any meaningful enquiries, yet software that is highly scalable and fast may be too rigid or opaque to quickly develop and test new ideas. There remains a need for tools that strike a fine balance between these two extremes for the purposes of conducting advanced research at scale.

Mava is a research library that aims to find such a balance, for online MARL research in particular. By leveraging the power of JAX as a machine learning framework~\citep{jax2023github}, along with advances in distributed computing~\citep{hessel2021podracer}, Mava is able to train over millions of timesteps in a matter of minutes, enabling highly efficient experiment iteration. Yet, simultaneously, by intentionally adopting and adhering to the best practices of simple and readable code, Mava remains easy to parse, debug, and extend.

\begin{figure}[t]
    \centering
    \includegraphics[width=\linewidth]{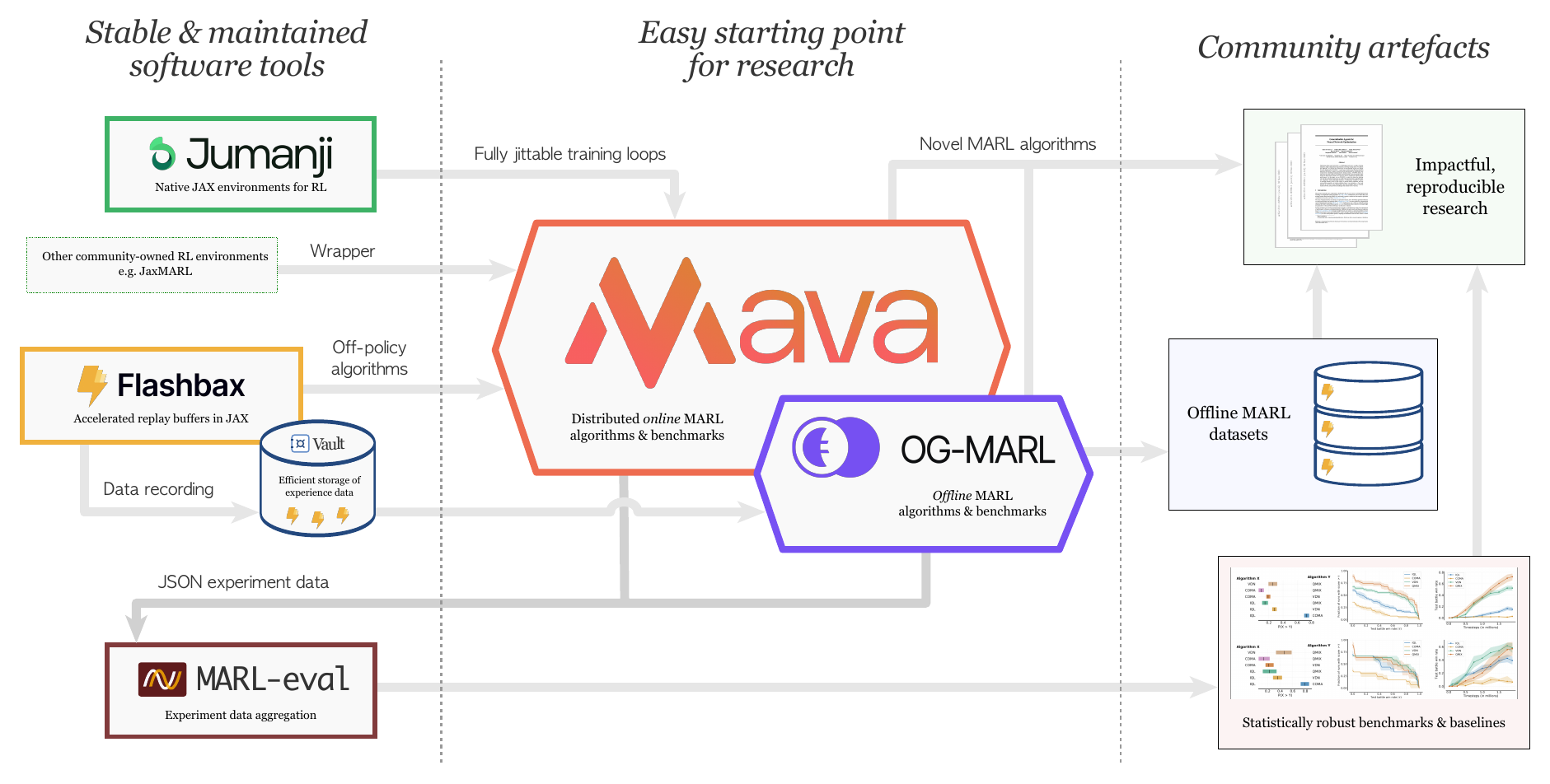}
    \caption{Depiction of how \href{https://github.com/instadeepai/mava}{\mava} fits into a broader multi-agent reinforcement learning ecosystem. Other components include \href{https://github.com/instadeepai/og-marl}{\ogmarl}, \href{https://github.com/instadeepai/jumanji}{\jumanji}, \href{https://github.com/instadeepai/flashbax}{\flashbax}, and \href{https://github.com/instadeepai/marl-eval}{\marleval}. Further details for each of these components are given in Section~\ref{sect:ecosystem}.}
    \label{fig:ecosystem}
\end{figure}

Importantly, Mava forms part of a broader ecosystem of research for MARL. Acting as an easy starting point for research ideas, Mava integrates naturally with a range of software libraries---spanning a corresponding framework for \emph{offline} MARL~\citep{formanek2023ogmarl} with accelerated replay buffers for off-policy and offline algorithms~\citep{flashbax2023github}, native RL environments written in JAX~\citep{bonnet2023jumanji}, and a robust and statistically reliable set of evaluation tools designed specifically for MARL~\citep{gorsane2022towards}. This ecosystem is depicted in Figure~\ref{fig:ecosystem}.

In this brief technical report, we provide an overview of the design and core features of Mava, and how it integrates into the broader MARL research ecosystem. Moreover, we conduct basic experiments demonstrating Mava's speed and performance on various tasks.

\section{Design and core features}

\textbf{A clean code philosophy.} Moving away from completely modular MARL frameworks \citep{samvelyan19smac, papoudakis2021benchmarking, hu2022marllib, bettini2023benchmarl}, Mava instead adopts a code philosophy akin to recent works such as CleanRL \citep{huang2022cleanrl} and PureJAXRL \citep{lu2022discovered} where all core algorithmic logic should be contained in a single easy-to-read file. This enables researchers to debug code easily and to adapt Mava to their particular use case with little friction and overhead. A notable difference between CleanRL and Mava is that Mava has certain levels of abstraction; on environment initialisation, reusable type definitions and flexible algorithm and environment configuration management with Hydra \citep{Yadan2019Hydra}. This approach strikes a balance where code is easy to understand but unnecessary boilerplate code is abstracted away.

\begin{figure}[htbp]
    \centering
    \includegraphics[width=\linewidth]{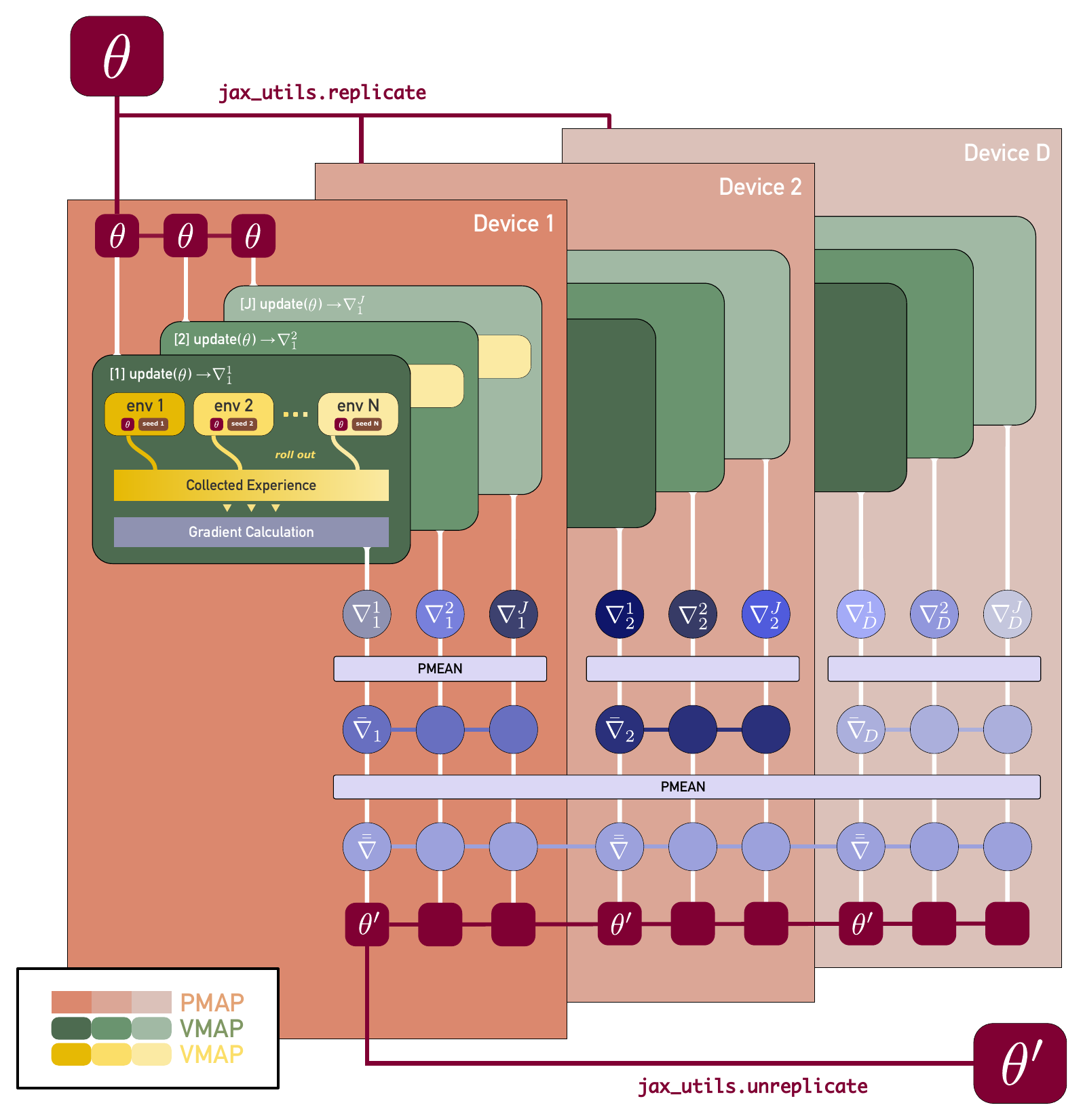}
    \caption{A diagram of the Anakin podracer architecture~\citep{hessel2021podracer} implemented in Mava for fast training on hardware accelerators, such as a GPU or TPU. A set of parameters $\theta$ is taken as input, and is replicated across $D$ devices. On each device $d \in D$ (via \texttt{jax.pmap}), these parameters are further broadcast to $J$ \texttt{update} functions (via \texttt{jax.vmap}). Each update function $j \in J$ rolls out independent experience from $N$ copies of the environment (via \texttt{jax.vmap}), which is collected and given to a loss function, for which a gradient is computed. For device $d$ and update function $j$, the gradient is $\nabla_d^j$. On each device $d$, the gradients of the $J$ update functions are averaged via a \texttt{jax.pmean} operation, yielding $\bar{\nabla}_d$. These gradients are then averaged across devices, again via \texttt{jax.pmean}, yielding $\bar{\bar{\nabla}}$, which is used to calculate the new parameters, $\theta'$.}
    \label{fig:mava-anakin}
\end{figure}

\textbf{Strong baseline algorithms and training architectures.} Mava currently only supports environments that are written in JAX. This constraint enables the (just-in-time) \texttt{jit}-compilation of agent policy roll-outs and policy updates. In particular, Mava supports the Anakin architecture~\citep{hessel2021podracer} for scalable distributed system training on hardware accelerators, depicted in Figure~\ref{fig:mava-anakin}. At the time of writing, Mava has implementations of both recurrent and feedforward policy versions of multi-agent PPO systems that follow the decentralised training with decentralised execution (DTDE) and centralised training with decentralised execution (CTDE) paradigms. Next in our roadmap is to include off-policy algorithms leveraging the Anakin architecture as well as both on and off-policy algorithms supporting distributed non-JAX environment training using the Sebulba architecture~\citep{hessel2021podracer}. 

\textbf{Multi-device training with easy checkpointing and logging.} Mava offers some core benefits over recent MARL offerings in JAX \citep{rutherford2023jaxmarl}. The first of which is out-of-the-box support for code to be \texttt{pmap}-ed over multiple devices on hardware accelerators like TPUs for faster training times. The second is support for robust evaluation, system checkpointing and continual metric logging. It is common in MARL systems to freeze training periodically in order to evaluate current agent policies \citep{gorsane2022towards}. This is a challenge in end-to-end JAX-based RL since \texttt{jit}-compiling the agent training loop forces one to wait for training to be complete before any performance feedback can be obtained. Possible ways around this are to either (1) use JAX's \texttt{debug} callback functionality, which tends to lead to significant increases in training wallclock time, (2) to evaluate systems once at the end of training, or (3) to forgo evaluation feedback altogether and only produce training curves - none of which is desirable. Mava solves this problem by interleaving evaluation blocks during system training. Here, both the training and evaluation blocks are \texttt{pmap}-ed and run sequentially in a normal Python for loop. This has the benefit that it maintains system performance while giving researchers continual insights into system training dynamics. Since system performance and parameters are continually exposed, metrics can be logged and policy parameters can be checkpointed continually throughout training. Mava has native support for logging to Tensorboard and Neptune and also supports logging to JSON files in a format that is supported by the MARL-eval \citep{gorsane2022towards} library for downstream statistical aggregation and plotting, discussed further below. 

\section{Seamless integration between ecosystem libraries \label{sect:ecosystem}}

As shown in \autoref{fig:ecosystem}, Mava forms part of a wider and evolving ecosystem of software for RL research. By design, this ecosystem allows for seamless integration between libraries.

\textbf{Accelerated JAX-native environments.}
One of the main challenges in RL and MARL is that algorithms tend to be sample inefficient, typically requiring millions of timesteps to converge. This means that the speed at which the environment can step has a substantial impact on the time to convergence of algorithms. Recently, environments implemented directly in JAX have demonstrated significant speedups in environment interactions per second by effectively leveraging modern hardware accelerators such as GPUs and TPUs \citep{gymnax2022github,brax2021github}. To take full advantage of this trend, Mava currently supports three suites of JAX-native multi-agent environments, namely \textbf{Matrax} \citep{matrax2023github}, \textbf{Jumanji} \citep{bonnet2023jumanji} and \textbf{JaxMARL} \citep{rutherford2023jaxmarl}. 

\begin{itemize}
    \item \textbf{Matrax} is a lightweight suite of common 2-player matrix games written in JAX. Matrix games are appealing for their simplicity and speed. As such, they provide a good starting point for testing new MARL algorithms and for discovering fundamental failure modes, even in simple settings. 
    \item \textbf{Jumanji} provides a suite of environments related to combinatorial optimisation problems. Several environments in Jumanji lend themselves well to being solved by MARL algorithms, for example, Multi-Robot Warehouse (RWARE) \citep{christianos2020shared}, Level-Based Foraging (LBF) \citep{albrecht2015game} and the Multi-Agent Capacitated Vehicle Routing Problem (MA-CVRP).  In Mava, we have developed flexible wrappers for Jumanji to make it simple to train Mava systems in these environments. 
    \item \textbf{JaxMARL} contains JAX implementations of many popular MARL benchmark environments, such as the StarCraft Multi-Agent Challenge (SMAC) \citep{samvelyan2019starcraft}, Overcooked \citep{carroll2019utility} and Multi-Agent MuJoCo (MAMuJoCo) \citep{peng2021facmac}. JaxMARL's environment offering is appealing because of its close relationship with well-established non-JAX environments. However, while it may be tempting to directly compare results on JaxMARL environments to their non-JAX versions, it's important to bear in mind that in some cases there can be subtle but important differences which make such comparisons invalid.
\end{itemize} 

As the ecosystem evolves, we plan to add support for additional environments. For researchers who are interested in using Mava on any currently unsupported environment, it should be quite easy to develop an environment wrapper similar to those for Jumanji or JaxMARL to make their new environment conform to the API Mava algorithms expect.

\textbf{Standardised and statistically robust evaluation reporting.} To ensure a high level of standardisation and statistical rigour in the evaluation of experimental outcomes, Mava can log raw experimental data in such a way that it conforms to the format expected by \textbf{MARL-eval}~\citep{gorsane2022towards}. MARL-eval implements the data aggregation and reporting proposed by \cite{gorsane2022towards}, based on the principles established in \citep{agarwal2021deep}. This equips MARL-eval with statistically sound plotting tools, enabling robust and reliable analysis of cooperative MARL experiments.

\begin{figure}
    \centering
    \includegraphics[width=0.4\linewidth]{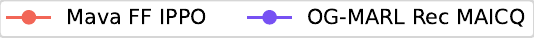}
    \includegraphics[width=0.98\textwidth]{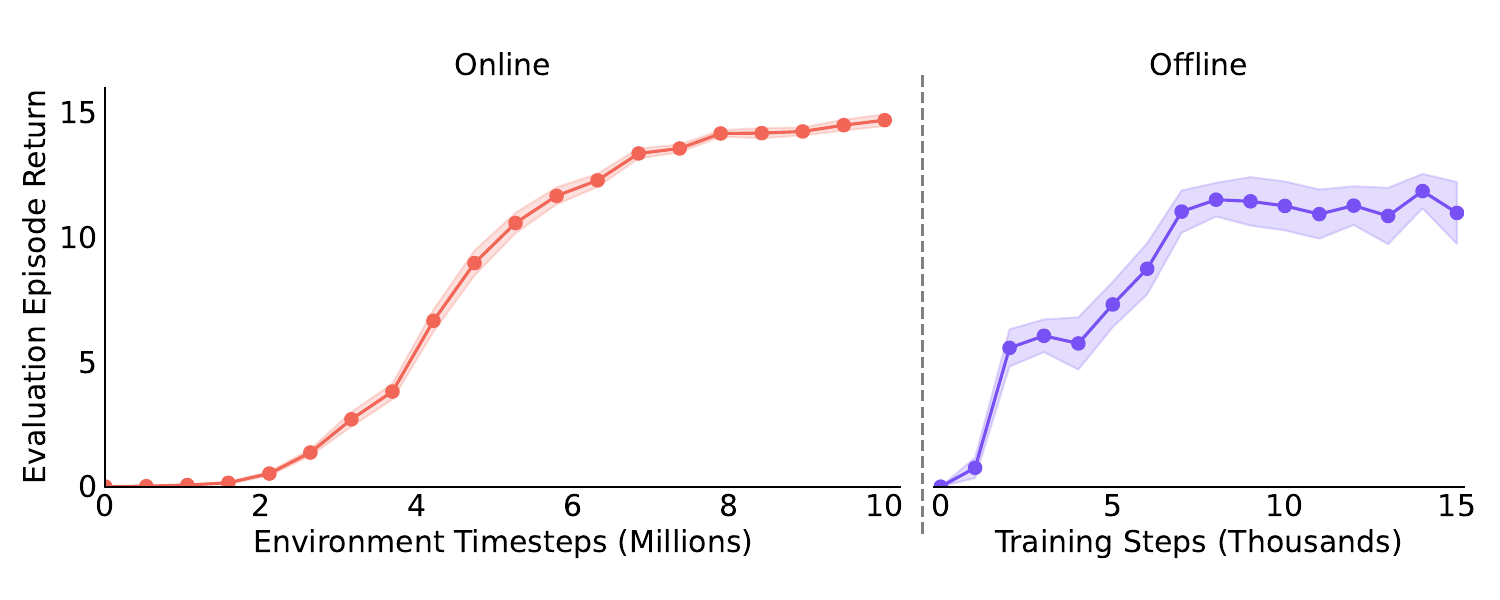}
    \caption{As a minimal demonstration of how Mava can be integrated into an offline MARL experiment pipeline, we trained an online Mava PPO system on RWARE and recorded its experience using a Flashbax Vault. We then reloaded the Vault in OG-MARL and trained the MAICQ algorithm completely offline on the Mava data. The entire online-to-offline experiment was repeated for 10 independent seeds. The mean and standard error across seeds are plotted.}
    \label{fig:online-to-offline-marl}
\end{figure}

\textbf{Offline MARL.}
A promising and increasingly popular research direction in MARL is offline training \citep{yang2021believe,tseng2022knowledge,meng2023offline,tian2023learning,wang2023offline}. In offline MARL, agents are trained on a static dataset of experience without any additional online interactions in the environment. Offline MARL is an appealing paradigm because it avoids online environment interactions which can be slow, expensive and dangerous if a simulator is not readily available. Another promising approach is combining offline with online training to make online training significantly more sample efficient\citep{nair2020awac,wagenmaker2023leveraging,ball2023efficient}. \citet{formanek2023reduce} demonstrated that a similar approach can significantly speed up online training in MARL. 

Recently, \textbf{Off-the-Grid MARL} \citep{formanek2023ogmarl} was proposed as a framework for offline MARL research. To demonstrate how Mava can be utilised in an offline research workflow, we recorded online transitions from a Mava PPO system on RWARE using a Flashbax Vault (see below) and reloaded these transitions as a dataset (directly into VRAM) to train a fully \texttt{jit}-compilable JAX version of the offline algorithm MAICQ \citep{yang2021believe} implemented in OG-MARL. A plot of the online and offline training curves is given in \autoref{fig:online-to-offline-marl}. 

\textbf{Flashbax Vault}~\citep{flashbax2023github} serves as the intermediary between Mava and OG-MARL. Vault offers an efficient mechanism to save Flashbax buffers, which essentially hold JAX arrays, to persistent data storage. Consider a Flashbax buffer which has experience data of dimensionality $(B, T, *E)$, where $B$ is a batch dimension (for the sake of recording independent trajectories synchronously), $T$ is a temporal dimension, and $*E$ indicates the multiple dimensions of the experience data. Since large quantities of data may be generated for a given environment, Vault extends the $T$ dimension to a virtually unconstrained degree by reading and writing \emph{slices} of buffers along this time axis. In doing so, gigantic buffer stores can reside on disk, from which sub-buffers can be loaded into RAM/VRAM for efficient offline training.

\section{Speed and Performance}

This section outlines a series of early experiments conducted to demonstrate the use and effectiveness of Mava. We stress that these experiments are not meant to serve as an extensive investigation, and in several cases, we suspect significant improvements can still be obtained through proper hyperparameter tuning. Nonetheless, these simple experiments show how Mava scales with respect to the number of vectorized environments leading to a significant reduction in experiment wallclock time (from several hours to only a few minutes), while simultaneously maintaining good performance.

We first compare Mava's speed and performance with a popular PyTorch-based MARL framework EPyMARL \citep{papoudakis2021benchmarking}, which itself is an extension of PyMARL~\citep{samvelyan19smac}, arguably one of the most popular frameworks in MARL. We conduct experiments in various Level-Based Foraging (LBF) \citep{albrecht2015game} and Multi-Robotic Warehouse (RWARE) \citep{papoudakis2021benchmarking} scenarios, utilizing both feedforward and recurrent architectures for Independent PPO (IPPO) and Multi-Agent PPO (MAPPO) with a centralised critic. Following these experiments, we provide a very preliminary comparison between Mava and the recently released JAX-based PPO baselines from JaxMARL on their StarCraft Multi-Agent Challenge in JAX (SMAX) environment  \citep{rutherford2023jaxmarl}. SMAX represents a JAX-based simplification of the original SMAC environment~\citep{samvelyan19smac}, streamlining its complexity for more flexible and fast experimentation. In all our experiments, we follow the evaluation guideline as proposed by \cite{gorsane2022towards}.

\begin{figure}
  \centering
  \begin{subfigure}[t]{\textwidth}
    \includegraphics[width=\textwidth, valign=t]{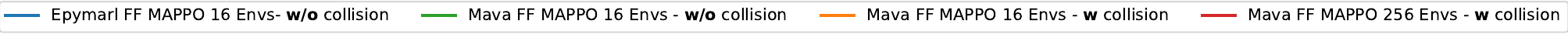}     
  \end{subfigure}
  \begin{subfigure}[t]{0.30\textwidth}
    \caption*{\hspace{1.5em} Tiny-2ag}
    \includegraphics[width=\textwidth, valign=t]{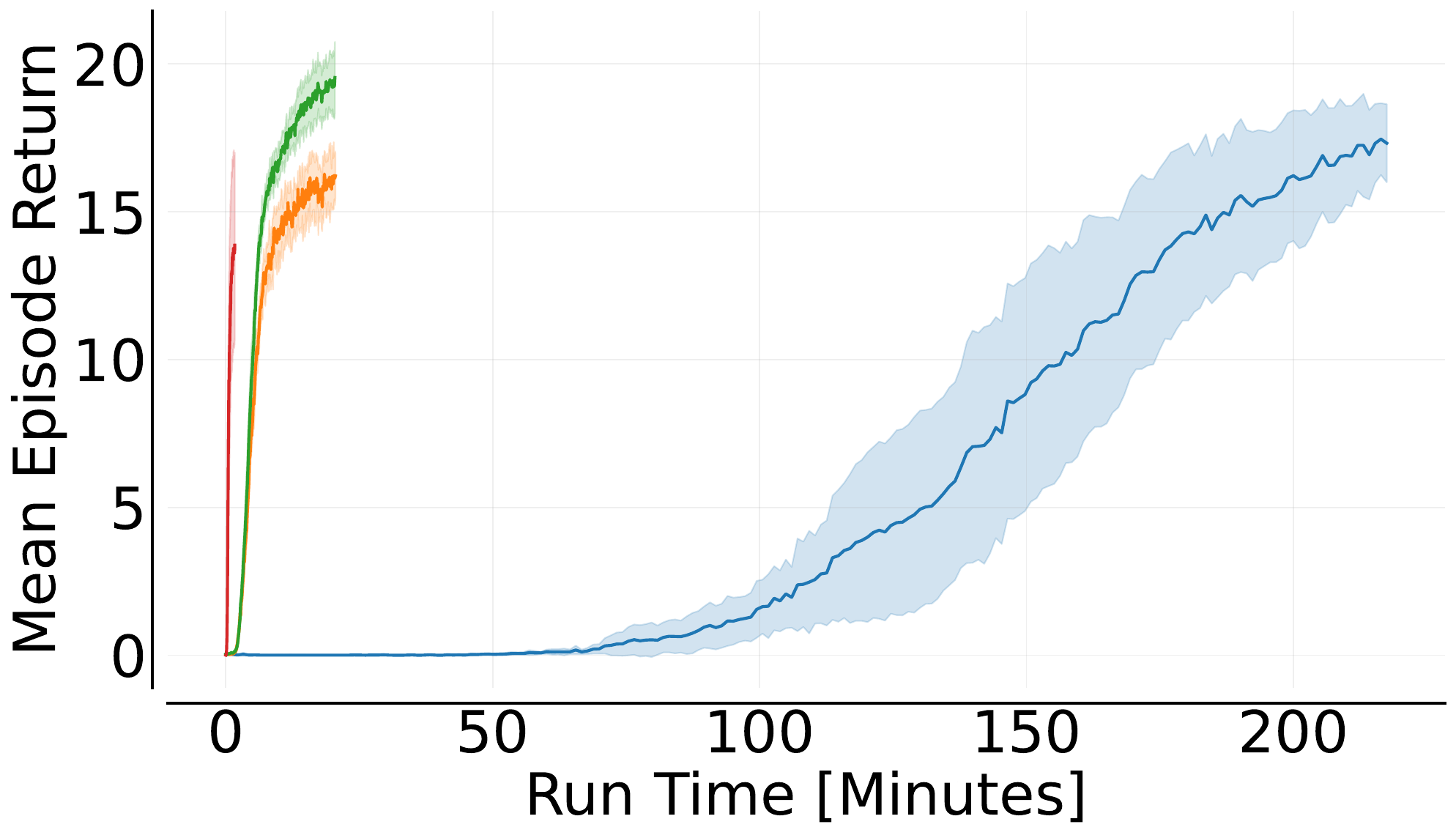}
    \label{fig:tiny-2ag}
  \end{subfigure}
  \hfill
  \begin{subfigure}[t]{0.30\textwidth}
    \caption*{\hspace{1.5em}Tiny-4ag}
    \includegraphics[width=\textwidth, valign=t]{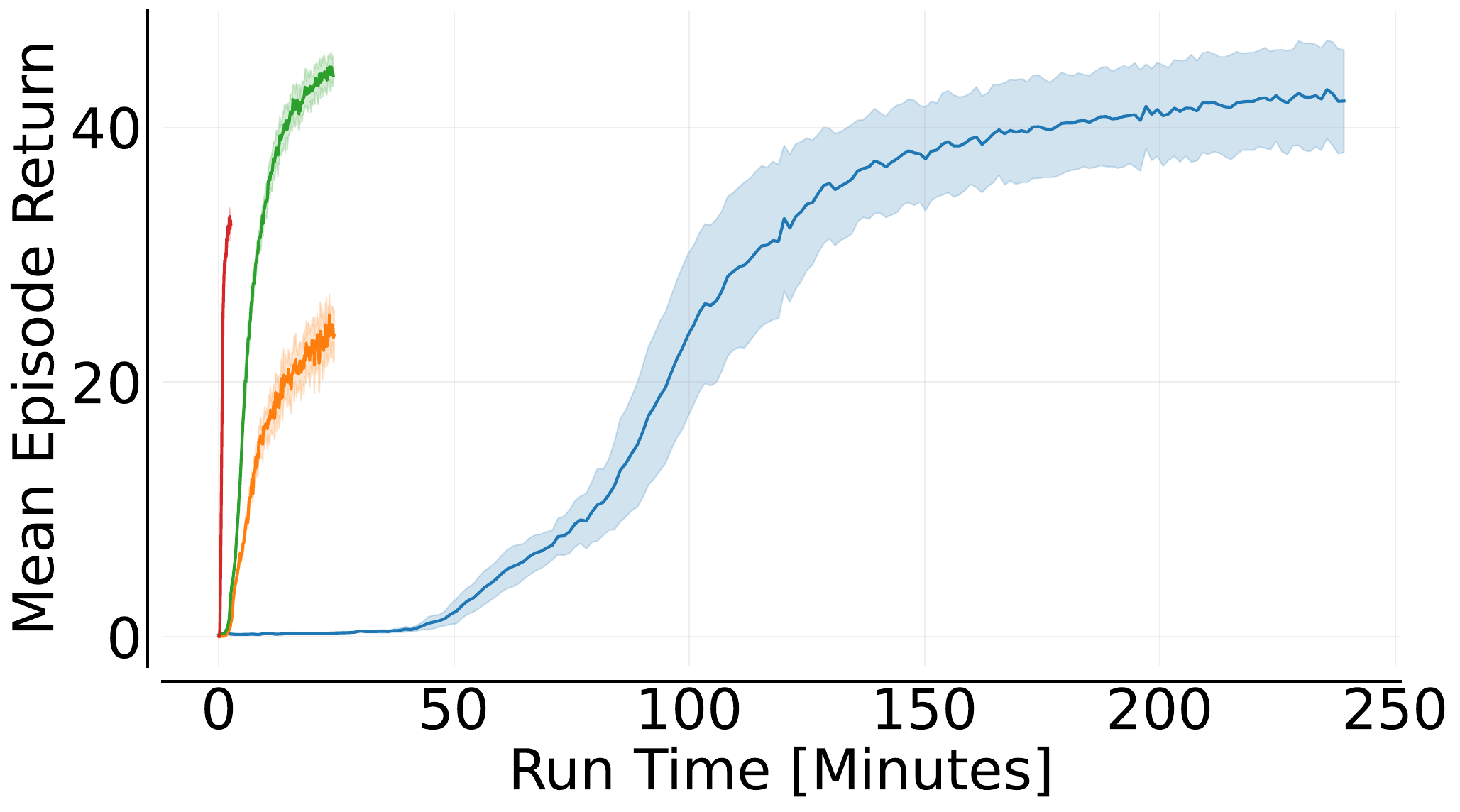}
    \label{fig:tiny-4ag}
  \end{subfigure}
  \hfill
  \begin{subfigure}[t]{0.30\textwidth}
    \caption*{\hspace{1.5em}Small-4ag}
    \includegraphics[width=\textwidth, valign=t]{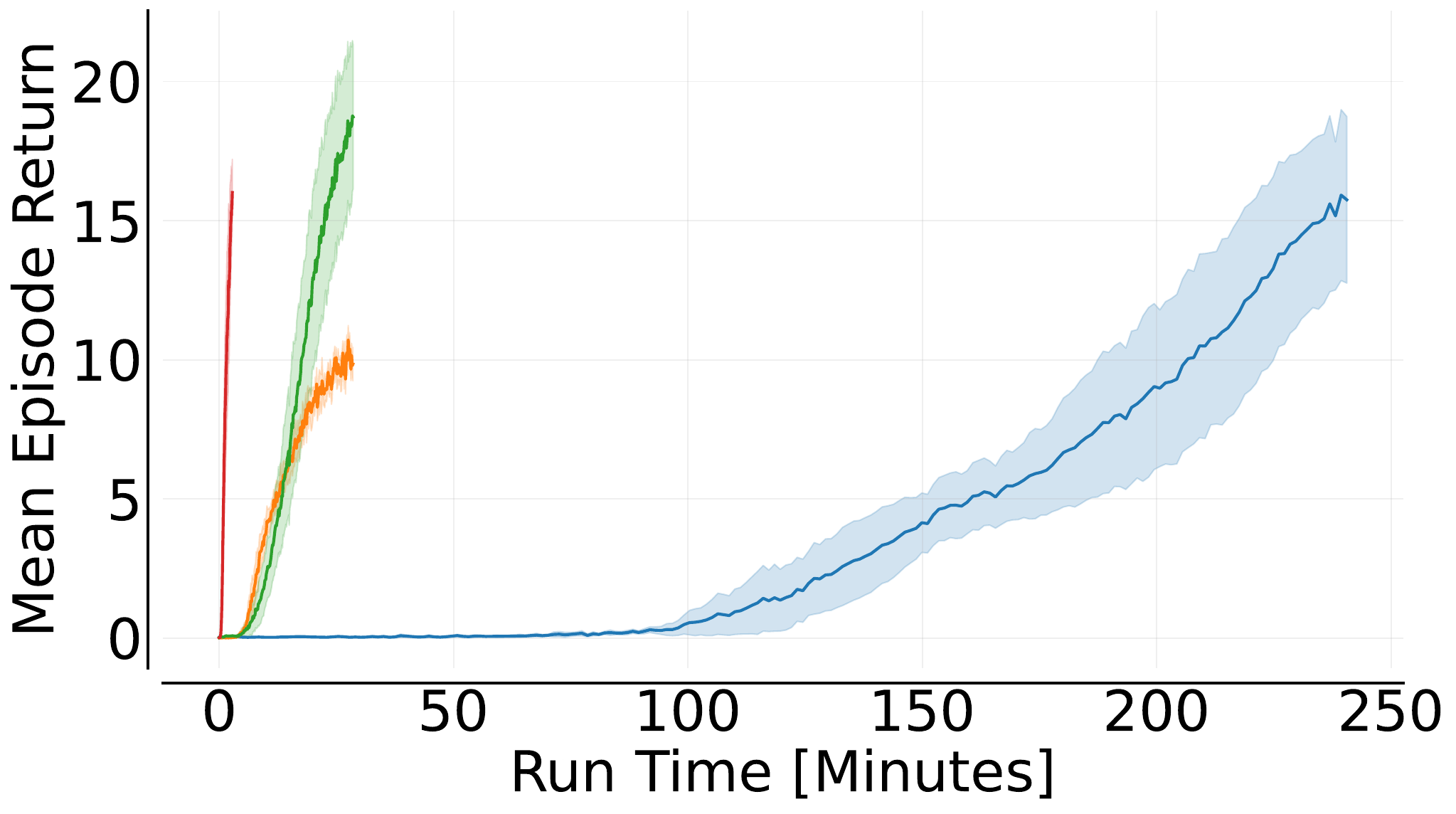}
    \label{fig:small-4ag}
  \end{subfigure}
  \caption{Performance comparison between Mava and EPyMARL training a feedforward MAPPO on the tiny-2ag, tiny-4ag, and  small-4ag tasks. Highlighting the comparable implementations between Jumanji's non-collision RWARE and the original RWARE, Mava's feedforward MAPPO shows equal or enhanced performance at significantly reduced wallclock time (from hours to minutes), with further speed gains observed when systems are scaled to use 256 \texttt{vmap}-ed achieving convergence in roughly 2 minutes.}
  \label{fig:ff-mappo-rware}
\end{figure}

\begin{figure}
    \centering
\begin{subfigure}[t]{0.02\textwidth}
\centering
        \scriptsize
        \textbf{(a)}
  \end{subfigure}
\begin{subfigure}[t]{0.45\textwidth}
\centering
    \includegraphics[width=\textwidth, valign=t]{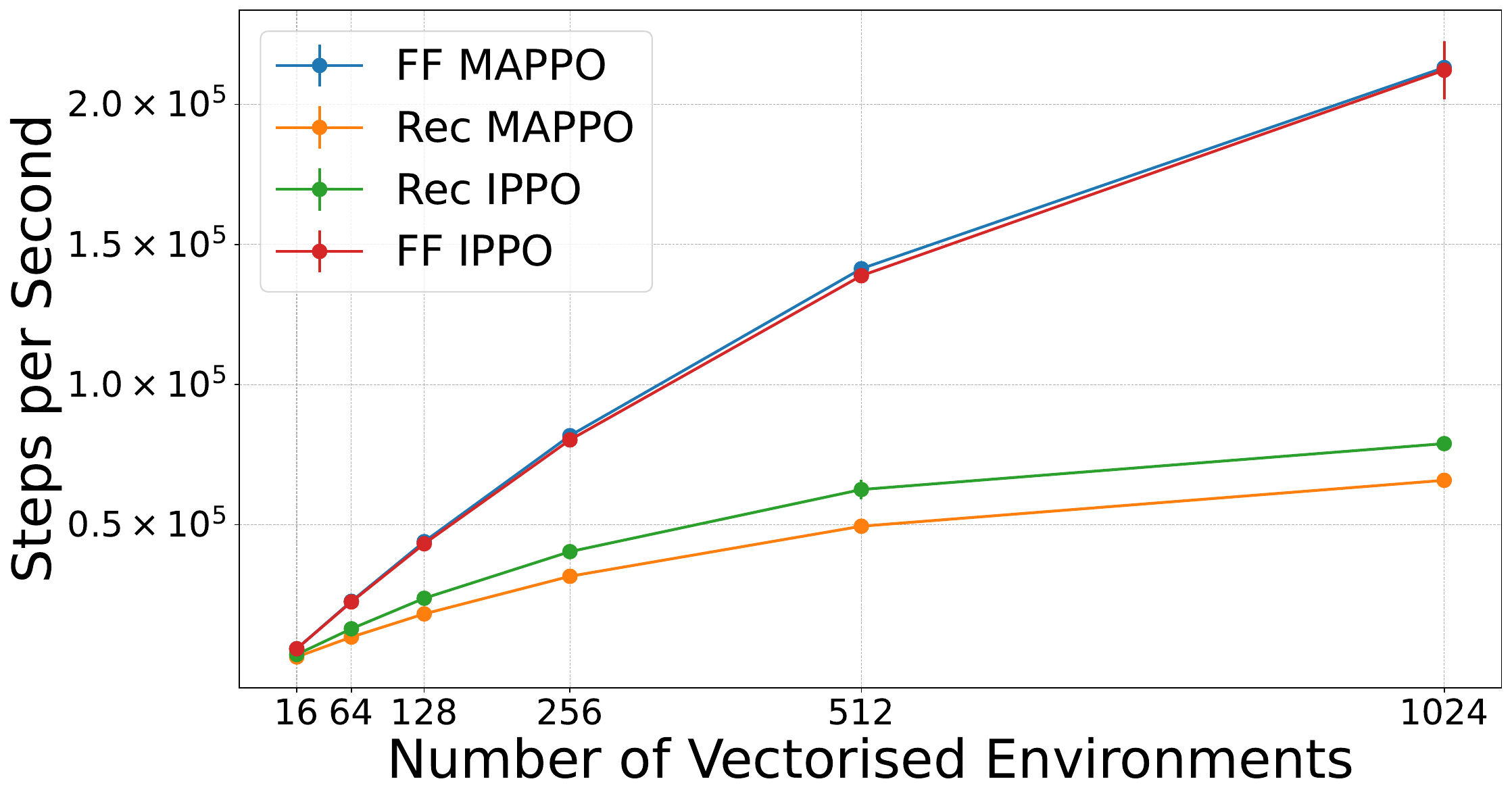}
    \label{fig:sps}
  \end{subfigure}
  \hspace{1em}
  \begin{subfigure}[t]{0.02\textwidth}
  \centering
        \scriptsize
        \textbf{(b)}
  \end{subfigure}
  \begin{subfigure}[t]{0.45\textwidth}
  \centering
    \includegraphics[width=\textwidth, valign=t]{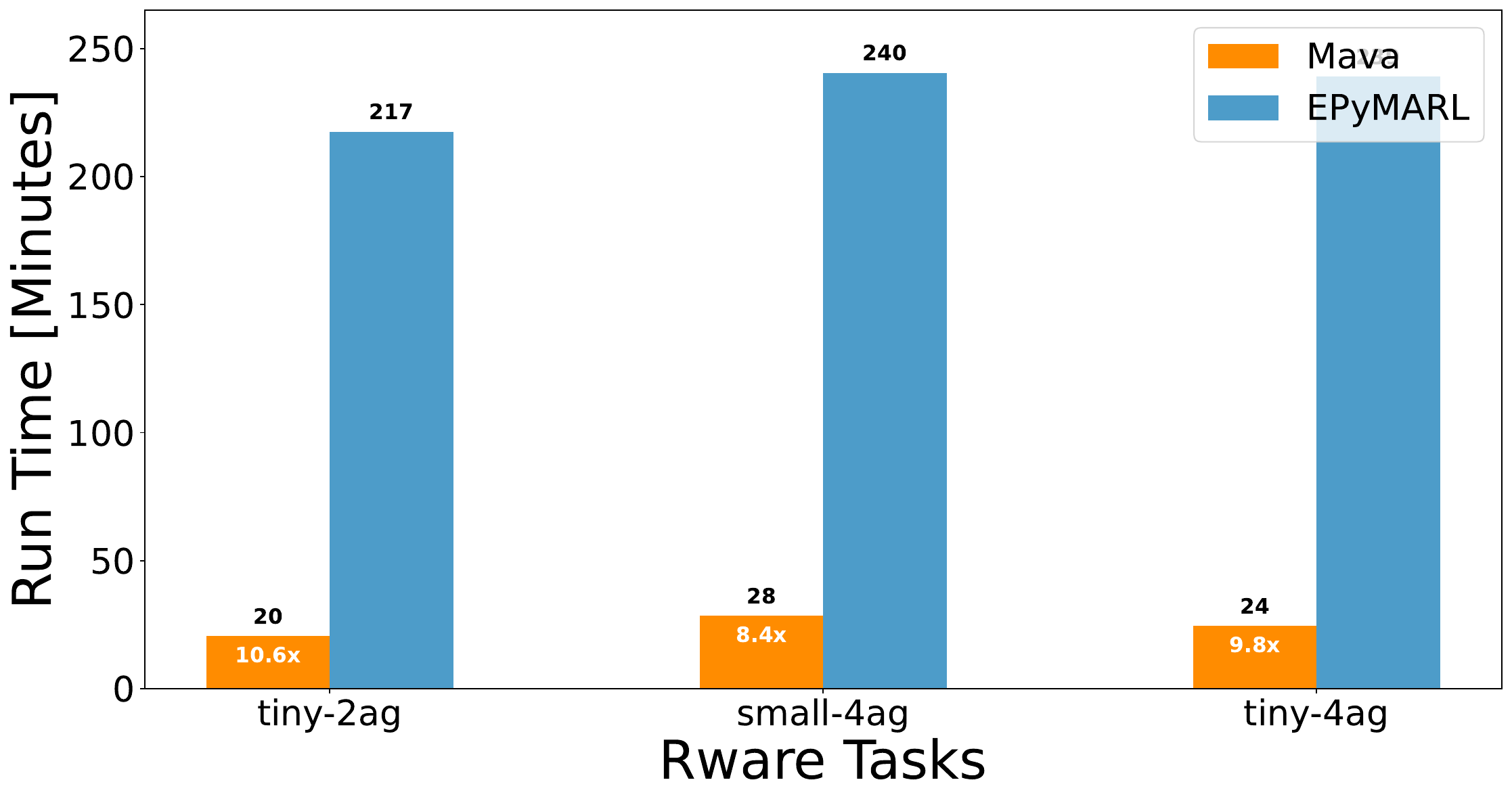}
    \label{fig:speed}
  \end{subfigure}
    \caption{Scalability and training time analysis: \textbf{(a)} illustrates the steps per second achieved by recurrent and feedforward IPPO and MAPPO as the number of vectorised environments increases, and \textbf{(b)} The bar chart compares the run time of Mava's implementations against EPyMARL's across the experiments from \autoref{fig:ff-mappo-rware} for the case of 16 \texttt{vmap}-ed environments, showcasing Mava's speed with about a 10x improvement across all tasks. At 256 vectorised environments we obtain improvements of more than 100x.}
    \label{fig:speed-mava}
\end{figure}

\textbf{Multi-Robot Warehouse (RWARE).} We train on 2 and 4-agent tasks, using Jumanji's RWARE for Mava and the original (non-JAX) RWARE for EPyMARL. We use the recommended hyperparameters from \citep{papoudakis2021benchmarking} for EPyMARL and conduct a basic grid search for Mava, training across 16 vectorized environments for up to 20 million steps. Notably, Jumanji's RWARE variant used in Mava terminates the episode on collisions, adding complexity; to ensure a fair comparison, we include Mava's performance with and without this feature. All experiments were conducted on an NVIDIA Quadro RTX 4000 GPU with 8GB Memory, leading to the showcased performance of FF-MAPPO in \autoref{fig:ff-mappo-rware} and the speed comparison in \autoref{fig:speed-mava}. Mava's speed is about 10x faster for the same number of parallel environments, and at 256 parallel environments, we obtain improvements of more than 100x in speed while maintaining good performance.

\begin{figure}
  \centering
  
  \begin{subfigure}[t]{0.9\textwidth}
    \includegraphics[width=\textwidth, valign=t]{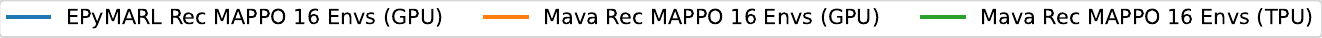}     
  \end{subfigure}

  \begin{subfigure}[t]{0.45\textwidth}
    \caption*{\hspace{2em}15x15-4p-3f}
    \includegraphics[width=\textwidth, valign=t]{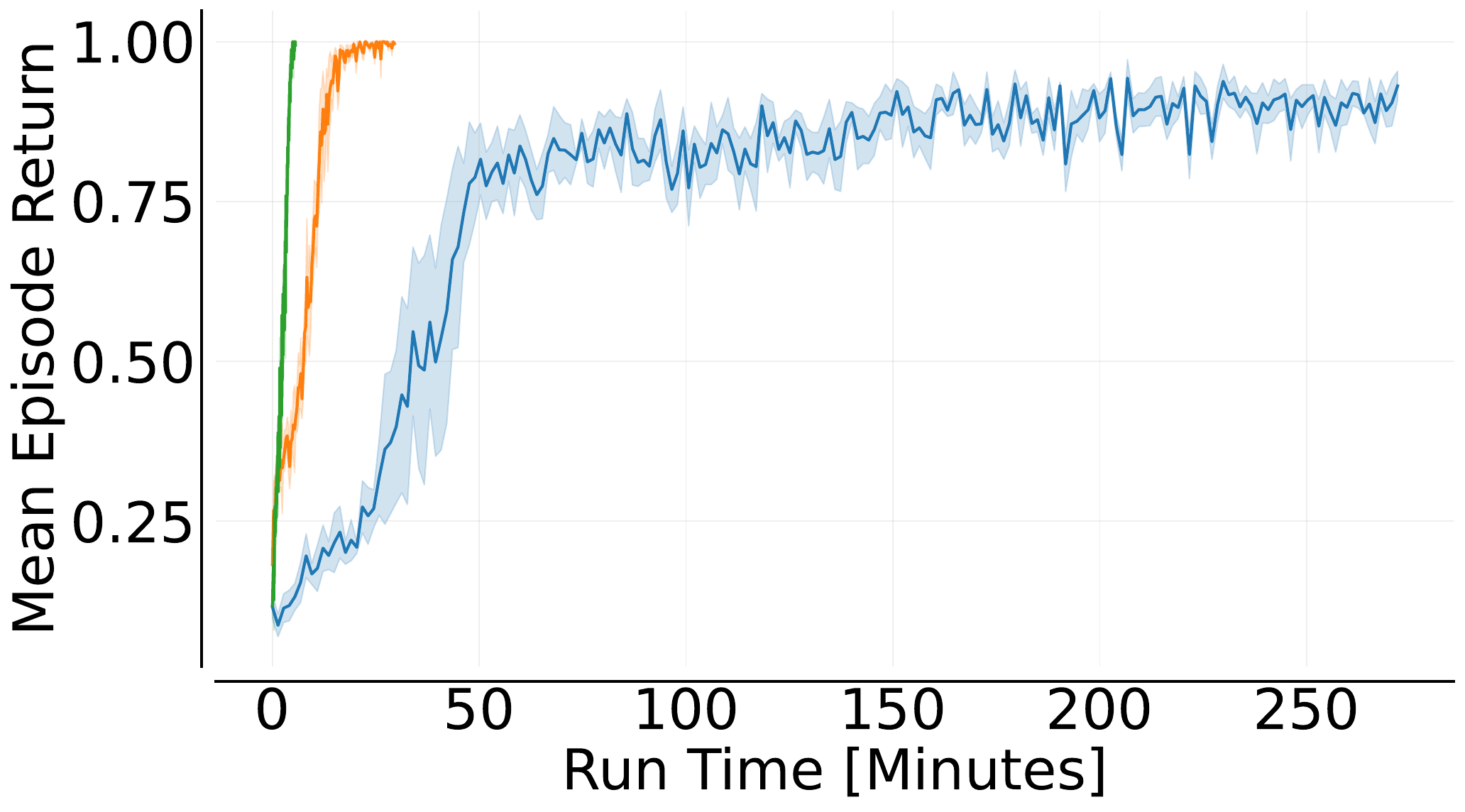}
    \label{fig:mappo-4ag}
  \end{subfigure}
  \hspace{1em}
  \begin{subfigure}[t]{0.45\textwidth}
    \caption*{\hspace{2em}2s-8x8-2p-2f}
    \includegraphics[width=\textwidth, valign=t]{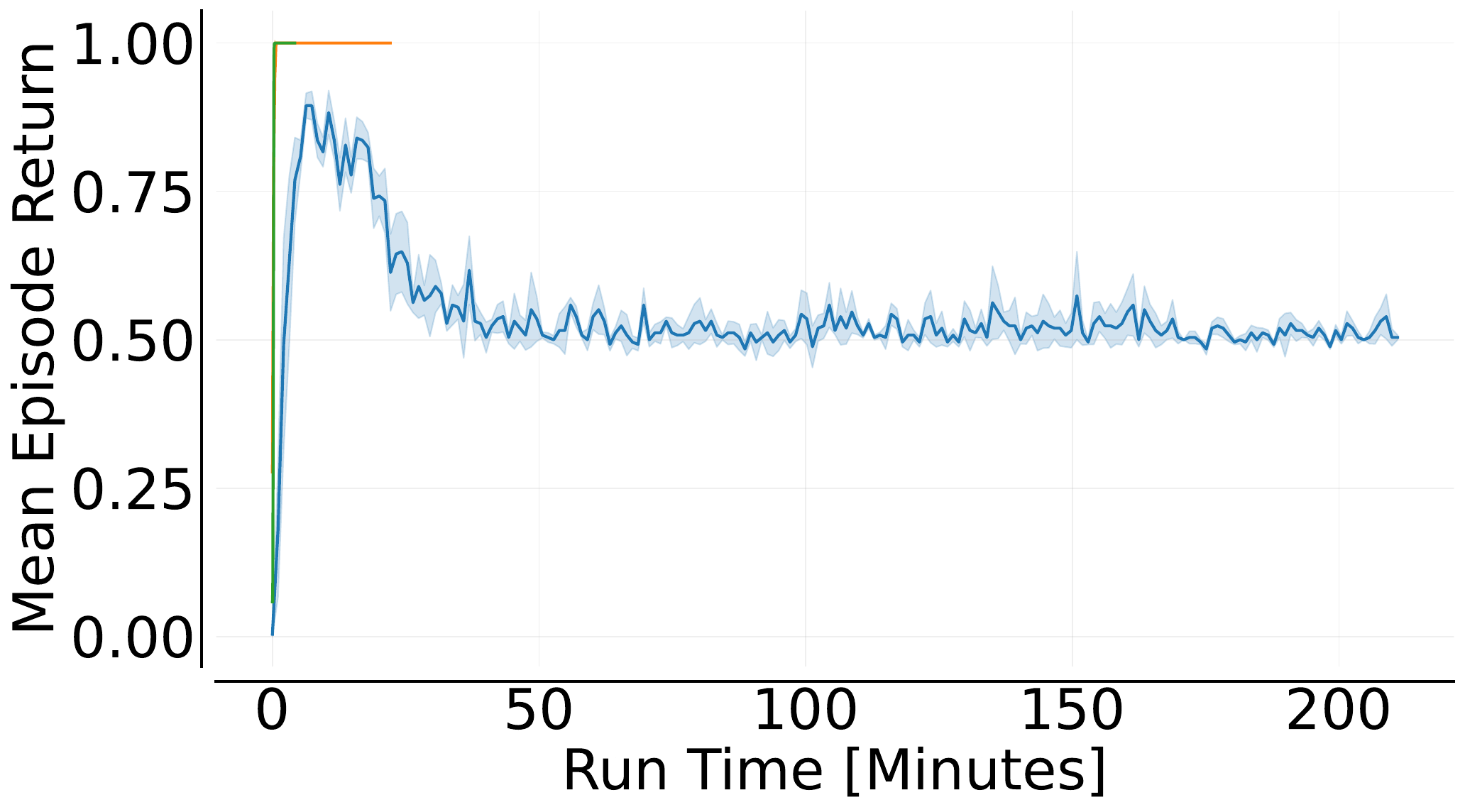}
    \label{fig:mappo-2ag}
  \end{subfigure}
  \caption{Performance comparison between Mava and EPyMARL training a recurrent MAPPO on the 15x15-4p-3f and 2s-8x8-2p-2f tasks. Mava achieves superior performance at 10x the speed. Moreover, by leveraging specialised hardware accelerators (in this case a TPU-V3), Mava is able to train to convergence in under five minutes.}
  \label{fig:lbf}
\end{figure}

\textbf{Level-Based Foraging (LBF).} We compare the performance of Mava's MAPPO recurrent system with those from EPyMARL in two settings: one involving 2 agents and the other with 4 agents, over 20 million timesteps. We utilize Jumanji's LBF for Mava and the original LBF for EPyMARL. Both setups are trained on 16 vectorized environments using a GPU (NVIDIA A-100 for EPyMARL and NVIDIA GeForce RTX 3050 with 4GB Memory for Mava). Additionally, Mava is tested with 16 parallel environments on a TPU-V3 to demonstrate scalability on specialised hardware. EPyMARL's hyperparameters are sourced from \citep{papoudakis2021benchmarking}, whereas we employ a basic grid search for Mava.

\begin{figure}
\centering
\includegraphics[width=0.8\linewidth]{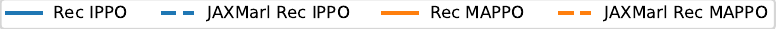}

\begin{subfigure}[t]{0.3\linewidth}
\caption*{\hspace{1.5em}2s3z}
\includegraphics[width=\linewidth, valign=t]{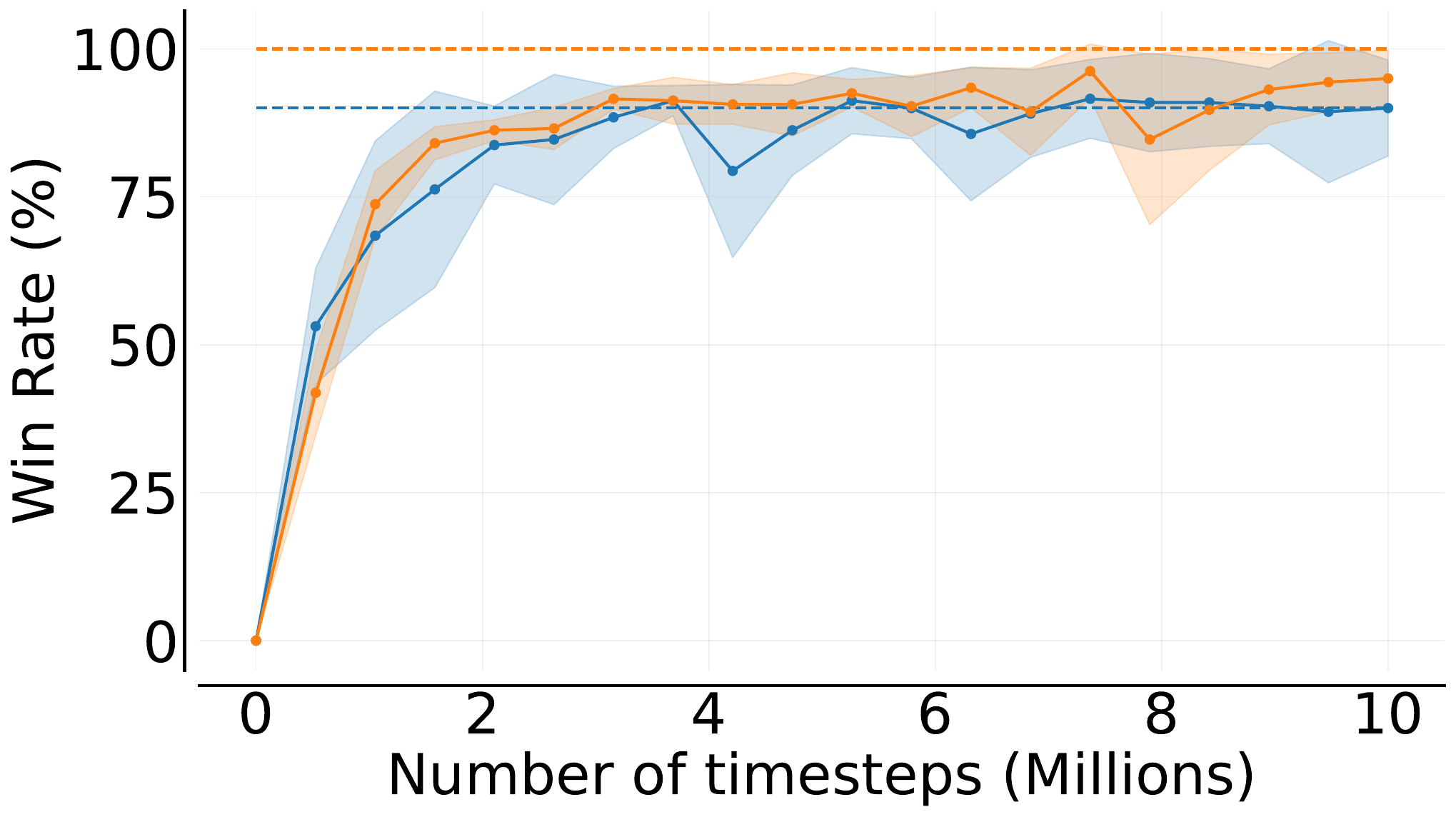}

\label{fig:2s3z}
\end{subfigure}
\hfill
\begin{subfigure}[t]{0.3\linewidth}
\caption*{\hspace{1.5em}3s5z}
\includegraphics[width=\linewidth, valign=t]{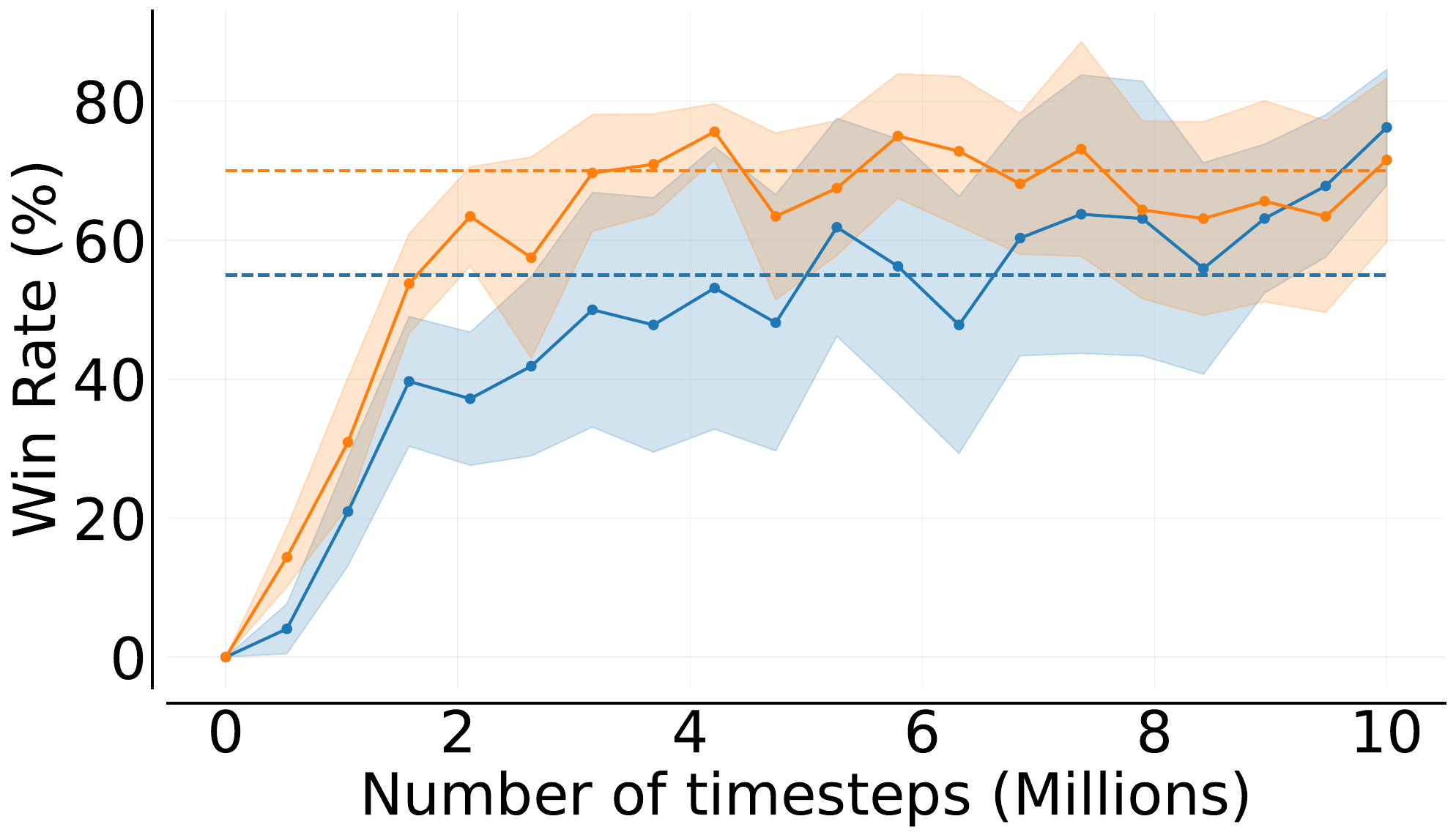}

\label{fig:3s5z}
\end{subfigure}
\hfill
\begin{subfigure}[t]{0.3\linewidth}
\caption*{\hspace{1.5em}6h vs 8z}
\includegraphics[width=\linewidth, valign=t]{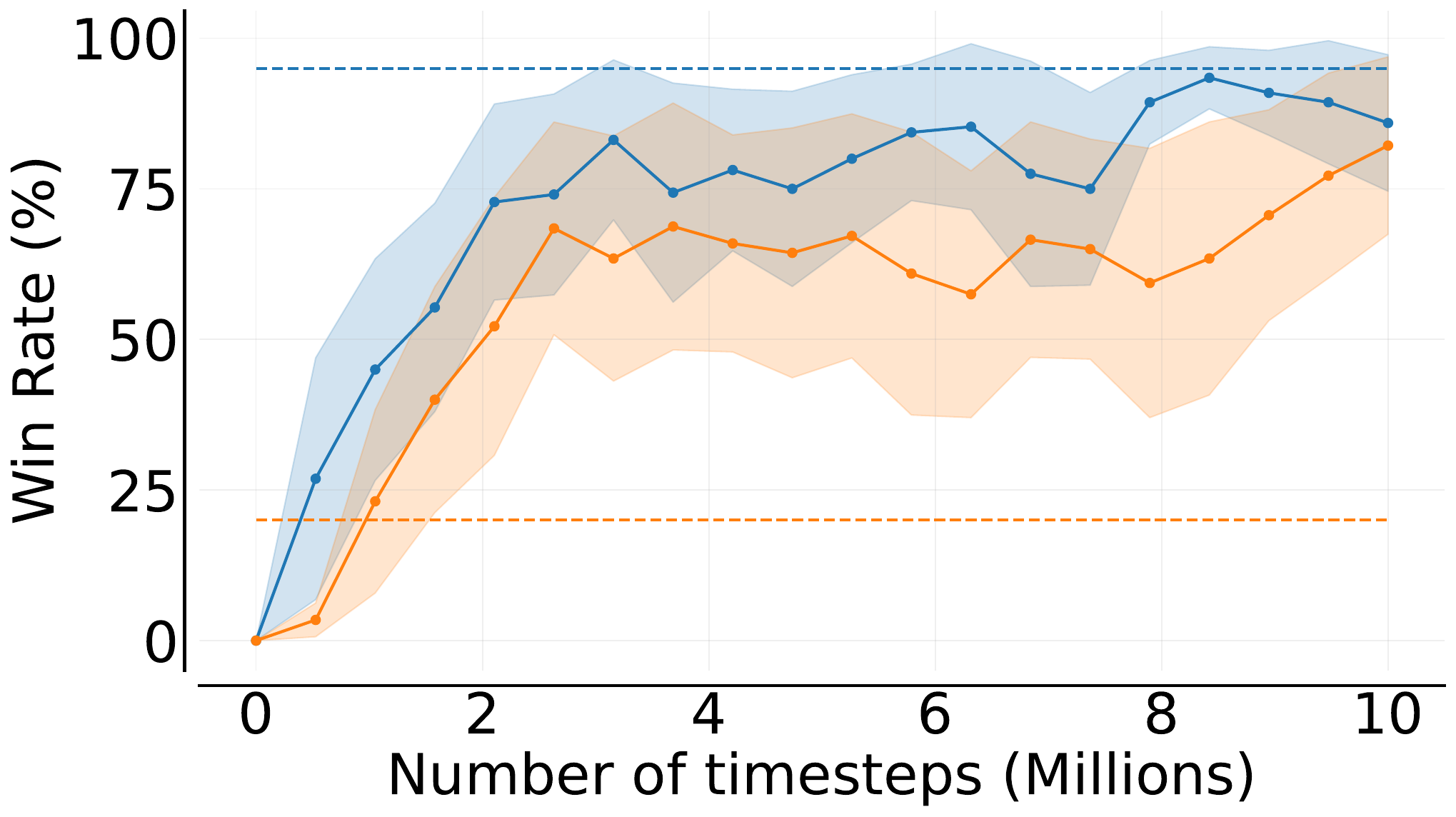}

\label{fig:6h_vs_8z}
\end{subfigure}

\caption{Performance of Recurrent IPPO and MAPPO from Mava and JaxMARL on SMAX Scenarios: 2s3z, 3s5z, and 6h vs 8z. Dashed lines indicate the estimated final win rates for JaxMARL taken from the plots in \citep{rutherford2023jaxmarl}. In all experiments, we performed no hyperparameter tuning for Mava systems but simply used the hyperparameters from JaxMARL.}
\label{fig:smax-tasks}
\end{figure}

\textbf{StarCraft Multi-Agent Challenge in JAX (SMAX).} We train Mava's recurrent systems on a similar subset of SMAX scenarios as used in the main text of \citep{rutherford2023jaxmarl}. We stress again that we do not consider these experiments to be extensive or sufficiently comprehensive to provide strong evidence to our claims, but only provide some preliminary results for interest. For all our experiments, we perform no hyperparameter tuning and simply use the hyperparameters from \citep{rutherford2023jaxmarl}, training for 10 million timesteps using 64 parallel environments. \autoref{fig:smax-tasks} illustrates that Mava's recurrent IPPO and MAPPO algorithms match the performance of JaxMARL's baselines. However, we suspect additional performance gains could be achieved with proper hyperparameter tuning. In terms of speed, we note that our FF-IPPO trains to convergence, achieving around 85\% win rate, on 2s3z in under a minute (54 seconds). From our interpretation of the speeds on 2s3z reported by \cite{rutherford2023jaxmarl} (in their Figure 5 (d)), the authors seem to claim single-run training times of roughly 5 seconds. However, we are unsure if we are interpreting these results correctly. In future we plan to run Mava (using both on and off-policy algorithms) across \textit{all} SMAX scenarios (the \textit{minimal} set suggested for research by \cite{rutherford2023jaxmarl} following the advice from \cite{gorsane2022towards}).

\section{Conclusion}
Mava is a flexible and performant JAX-based library for MARL that integrates seamlessly with several other powerful libraries forming a broader and continuously evolving ecosystem for MARL. By open-sourcing Mava we hope it can benefit the research community and help drive progress in the field.  

\newpage

\acks{We would like to thank Achraf Garai for helping with the logo designs. This research was supported with Cloud TPUs from Google's TPU Research Cloud (TRC).}


\noindent

\vskip 0.2in
\bibliography{main}

\newpage

\appendix

\end{document}